\documentclass[letterpaper,10pt,conference]{ieeeconf} 
\IEEEoverridecommandlockouts                            

\usepackage[noadjust]{cite}

\usepackage{xparse}
\usepackage{amsmath}
\usepackage{amssymb}
\usepackage{amsfonts} 
\usepackage{xcolor}
\usepackage{url}
\usepackage{balance}

\usepackage{graphicx}
\usepackage{caption}
\usepackage{subcaption}
\usepackage{etoolbox}

\makeatletter
\let\NAT@parse\undefined
\makeatother
\usepackage[colorlinks]{hyperref}
\usepackage{cleveref}
\crefformat{equation}{(#2#1#3)}
\Crefname{figure}{Fig.}{Figs.}

\usepackage{booktabs}
\usepackage[para]{threeparttable}
\usepackage{multirow}
\usepackage{makecell}
\usepackage{siunitx}

\title{\Large\bf PhotoBot: Reference-Guided Interactive Photography via Natural Language}

\author{Oliver Limoyo$^{\dag,1,2}$, Jimmy Li$^{\dag,1}$, Dmitriy Rivkin$^{1}$, Jonathan Kelly$^{2}$, and Gregory Dudek$^{1,3}$
\thanks{$^{\dag}$ Denotes equal contribution.}
\thanks{$^{1}$Authors are with the Samsung AI Centre, Montr\'{e}al, Qu\'{e}bec H3A 3G4, Canada. Email: o.limoyo@partner.samsung.com, \{jimmy.li, d.rivkin,steve.liu,greg.dudek\}@samsung.com}%
\thanks{$^{2}$Authors are with the STARS Laboratory, University of
Toronto Institute for Aerospace Studies, Toronto, Ontario M5S 1A4, Canada. Email: \{\textit{first-name.last-name}\}@robotics.utias.utoronto.ca}%
\thanks{$^{3}$Authors are with McGill University,  Montr\'{e}al, Qu\'{e}bec H3A 089, Canada. Email:  \{\textit{first-name.last-name}\}@mail.mcgill.ca}%
}

\begin{document}
\maketitle
\thispagestyle{empty}
\pagestyle{empty}
\begin{abstract}
We introduce PhotoBot, a framework for fully automated photo acquisition based on an interplay between high-level human language guidance and a robot photographer. 
We propose to communicate photography suggestions to the user via reference images that are selected from a curated gallery. 
We leverage a visual language model (VLM) and an object detector to characterize the reference images via textual descriptions and then use a large language model (LLM) to retrieve relevant reference images based on a user's language query through text-based reasoning.
To correspond the reference image and the observed scene, we exploit pre-trained features from a vision transformer capable of capturing semantic similarity across marked appearance variations.
Using these features, we compute suggested pose adjustments for an RGB-D camera by solving a perspective-n-point (PnP) problem.
We demonstrate our approach using a manipulator equipped with a wrist camera.
Our user studies show that photos taken by PhotoBot are often more aesthetically pleasing than those taken by users themselves, as measured by human feedback.
We also show that PhotoBot can generalize to other reference sources such as paintings.
\end{abstract}

\section{Introduction} 
\label{sec:int}

\begin{figure}
\vspace{1.5mm}
    \centering
    \includegraphics[width=\columnwidth]{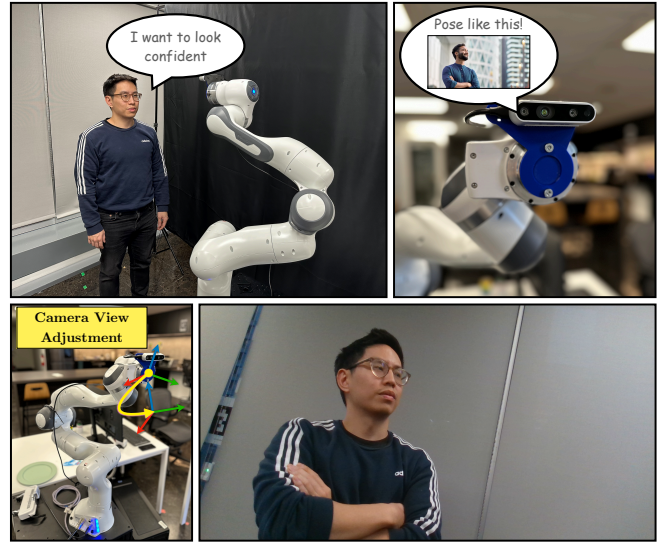}
    \caption{PhotoBot provides a reference photograph suggestion based on an observation of the scene and a user's input language query (upper left). The user strikes a pose matching that of the person in the reference photo (upper right) and PhotoBot adjusts its camera accordingly to faithfully capture the layout and composition of the reference image (lower left). The lower-right panel shows an unretouched photograph produced by PhotoBot.}
\label{fig:front_page}
\vspace{-2.5mm}
\end{figure}

Photographing a human subject requires nuanced interaction and clear communication between the photographer and the model.
Beyond just capturing well-composed photos, a professional photographer needs to understand what the client wants and to provide suggestions.
Much of the prior research in the area of robotic photography \cite{6907096, 1249268, 1570635, saycheese} has focused on the technical aspects, that is, how to navigate, plan, and control a robot to frame a photo, but not on the interaction between the photographer and human model.
Further, approaches relying on heuristic composition rules may not produce captivating photos, in part because quantifying the aesthetic quality of all images in a generalizable way is a difficult, unsolved problem \cite{6247954, 10.1007/11744078_23}.

\begin{figure*}[htb!]
    \includegraphics[width=0.98\textwidth]{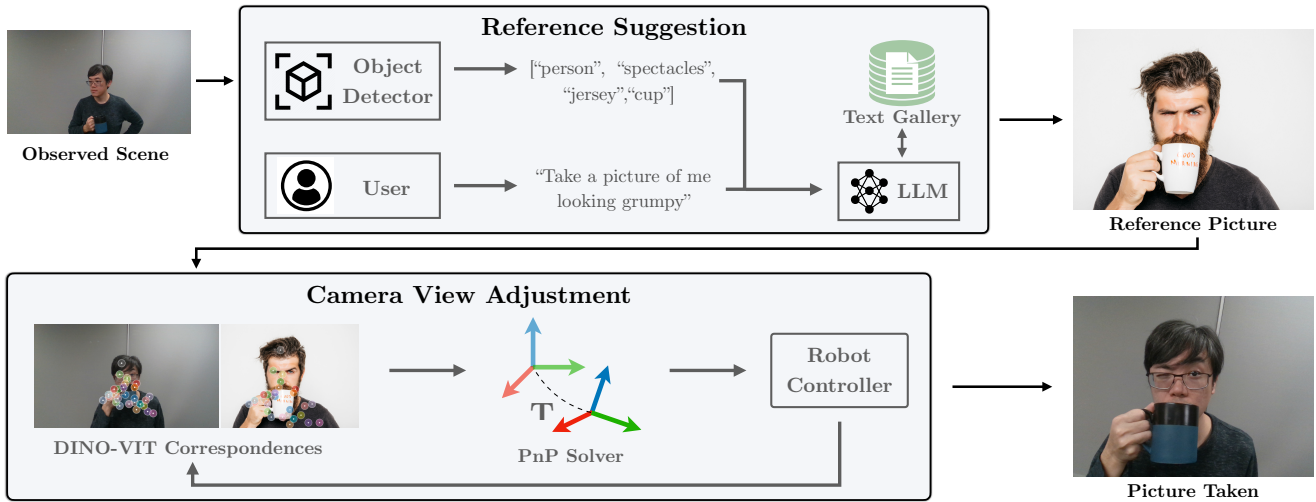}
    \centering
    \caption{PhotoBot system diagram. The two main modules are shown: Reference Suggestion and Camera View Adjustment. Given the observed scene and a user query, PhotoBot suggests a reference image to the user and adjusts the camera to take a photo with a similar layout and composition to the reference image.} 
\label{fig:system_diagram}
\vspace{-3mm}
\end{figure*}

In this work, we introduce PhotoBot, a language-guided assistant that provides photography suggestions based on high-level human guidance and computes camera view adjustments to produce aesthetically pleasing pictures. 
PhotoBot is capable of interacting with the user by leveraging the reasoning capabilities of large language models (LLMs) \cite{talmor2021commonsenseqa} and the grounding capabilities of visual language models (VLMs) together. 
Specifically, we first convert a curated gallery of images into text-based descriptions (e.g., including information such as a general description of the image, the mood, and the number of people in the image) using a VLM and an object detector. 
The VLM and object detector provide an automated approach to describe curated images using language.
Given a language query from a user and the detected objects in the scene observed by a camera, we use a LLM to retrieve a relevant \textit{reference image} (i.e., an existing image from the curated gallery of high-quality photographs) to suggest to the user through text-based reasoning.
The user then imitates what is shown in the image; PhotoBot solves for the respective camera motion and image crop such that the camera view matches the reference image. 
We formulate camera view adjustment as a perspective-n-point (PnP) problem \cite{10.1145/358669.358692} with pre-trained features from a vision transformer capable of capturing semantic similarity across significantly varying images \cite{amir2021deep, caron2021emerging}.

The learned components of PhotoBot operate together to yield a scaleable and generalizable system that can capture subjective and difficult-to-quantify aesthetic preferences through natural interaction.
\Cref{fig:front_page} shows an example of a problem instance and a photograph produced fully automatically by PhotoBot.
In summary, our main contributions are:
\begin{enumerate}
    \item the formulation of a new, imitation-based approach to photography using reference image templates;
    \item a novel grounded photo suggestion module building on a combination of a visual language model (VLM), an object detector, and a large language model (LLM);
    \item an experimental analysis of the perspective-n-point (PnP) problem with learned semantic-level keypoint correspondences across widely different scenes; and
    \item a user study evaluating the quality of photos taken by PhotoBot.
\end{enumerate}
 
\vspace{-1.25mm}
\section{Related Work} 
\label{sec:rw}

Robot photography has previously been studied in the context of mobile robotics. 
Early work \cite{1249268,saycheese} introduced a mobile robotic system that navigates, detects faces, and takes photographs based on hand-engineered composition rules. 
In \cite{5650341}, a combination of sound and skin detection is used to frame subjects.
An autonomous robotic photographer is introduced in \cite{6907096}, where head detection and handcrafted photography composition rules are applied with a subsumption control module to capture photos.
The KL divergence between the distribution of the facial positions and directions of photographed subjects and a target distribution following common composition rules is used to evaluate photography quality in \cite{article}.
Within the context of mobile robotics, much of the prior work has focused on the technical navigation and control problems and not on the social aspect (i.e., the interaction between the photographer and human subject) as tackled in our work. 

The authors of \cite{1570635} introduce a method to frame human and non-human subjects based on motion parallax and optical flow techniques in posed and cooperative settings. 
Methods such as \cite{10.5555/2283696.2283744} have also studied generalizing autonomous photography to scenes that do not contain human subjects, using various aesthetic criteria. 
Similar to \cite{1570635} and \cite{10.5555/2283696.2283744}, our framework generalizes to non-human subjects.
We also consider the posed and cooperative setting as done in \cite{1570635}.

Learning has been used in robot photography to formulate aesthetic models \cite{9341086} and, more directly, in approaches that apply reinforcement learning to find policies that optimize a reward function based on aesthetics \cite{alzayer2021autophoto}.

User or subject interactivity in the context of robot photography is addressed in \cite{4059330} by having the robot move towards users who are waving their hands. 
In our work, we are also interested in user interactivity. 
However unlike in \cite{4059330}, which focuses on attracting the attention of the robot photographer, we focus on a different task: suggesting photo ideas to the user.
Closer to our approach, \cite{10161403} uses a LLM to produce text descriptions of photos that a photographer would typically be expected to take at an event that is described at a high level by the user.
A VLM is then used to find the best image matches (according to an embedding distance) to these text descriptions from a video stream retrieved from a camera on a robot.
We instead focus on producing photo suggestions based on personal user queries and grounded scene observations. 
We provide reference images as suggestions and not textual descriptions of photos as done in \cite{10161403}.
Our searching and matching procedure is carried out purely in text space by using a LLM and text descriptions of existing images in a gallery, as opposed to using a VLM. 
The use of the LLM enables more sophisticated reasoning and better explainability.
In \cite{10354952}, the authors use an LLM to generate trajectories that capture specific videos and photographs based on user language queries.
The authors program various camera movement primitives ahead of time, which the LLM can then call.
Similar to our work, an object detector is used to ground the LLM in the real world. 
However, we leverage the LLM for photography curation and not for control.

The social dimension of photography has also been explored beyond the realm of natural language alone.
In \cite{10.1145/3319502.3374809}, humorous content is displayed by a robot photographer to elicit spontaneous smiles. 
In \cite{10.1145/3544548.3580896}, the authors design a system to autonomously capture how-to videos and related photographic content; the robot photographer detects body, hand, and text cues to determine the regions of interests to track. 
We focus on re-creating reference images, which is more challenging from a semantic content perspective but simultaneously allows us to simplify the camera view adjustment problem as a PnP problem.

\section{Method} 
\label{sec:me}

Our proposed pipeline consists of two modules: a reference suggestion component that retrieves reference images for the user to imitate and a camera view adjustment component that alters the camera view to achieve a similar photo composition to the reference. 
An overview of the system is shown in \Cref{fig:system_diagram}. 
We describe the overall user interaction workflow in \Cref{subsec:ui}, the reference image selection process in \Cref{subsec:ts}, the extraction of semantic correspondences in \Cref{subsec:semkpt}, viewpoint alignment in \Cref{subsec:vpalign}, and outlier removal in \Cref{subsec:outliers}.

\subsection{User Interaction Workflow}
\label{subsec:ui}

The steps in our proposed workflow for reference-guided photography are as follows.
\begin{enumerate}
\item The user issues a query (e.g., ``Take a picture of me looking happy.'').

\item PhotoBot detects objects and the number of people in the environment.

\item Taking both the user query, detected objects, and number of people into consideration, the system retrieves a shortlist of relevant reference images from a curated gallery.

\item The user selects a preferred reference image from the shortlist.

\item The user imitates what is shown in the selected reference image.

\item PhotoBot moves the camera such that the view matches that of the reference image.

\item PhotoBot captures the photo and crops it to match the reference image aspect ratio.
\end{enumerate}
In the sections below, we describe each component and step in more detail.

\subsection{Reference Suggestion}
\label{subsec:ts}

\begin{figure}[t]
\includegraphics[width=\columnwidth]{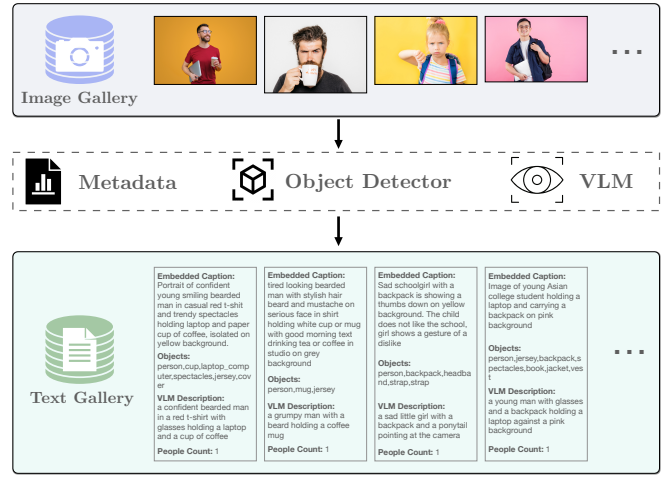}
\centering
\caption{We convert a gallery of curated reference images into a text-based representation using a combination of readily-available metadata, an object detector, and a VLM. A text-based gallery enables a LLM to search, match, and suggest reference images based on a language query from a user and a list of detected objects in the current scene.} 
\label{fig:gallery_conversion}
\vspace{0mm}
\end{figure}

\begin{figure}[t]
\includegraphics[width=\columnwidth]{figs/llm_query.pdf}
\centering
\caption{Examples of user queries and objects detected in the scene and the resulting reference image suggested by the LLM. We explicitly query the LLM to explain its choice of image suggestion. We can also query for an image suggestion without any information from the observed scene, as shown in the fourth row.}
\label{fig:llm_query}
\vspace{-4mm}
\end{figure}
To prepare the curated reference image gallery mentioned in Step 3 of \Cref{subsec:ui}, we obtain a set of high-quality, professionally-taken photos from the Internet. 
We preprocess each photo by querying a VLM to describe the image and count the number of people that appear. 
Additionally, we use an object detector to identify all of the objects that are present. 
If available, we also extract textual details embedded in the metadata of the photo.
We use Detic \cite{zhou2022detecting} as our object detector and InstructBLIP \cite{instructblip} as our VLM.
Finally, the VLM description, object list, metadata, and people count are concatenated into a single textual caption, which we embed into a vector using a sentence transformer \cite{reimers-2019-sentence-bert}.
We visualize this procedure in \Cref{fig:gallery_conversion}. 
The embedding vectors of all reference images are entered into a vector database for efficient and scalable retrieval.

At execution time, we create a textual embedding of the user prompt, which consists of the user's query from Step 1, detected objects in the current view, and the people count for the current view. 
We calculate the cosine similarities between the user prompt and all other vectors in the database and retrieve the top $m$ most similar image descriptions. 
Many of these image descriptions are only coarsely related to the user query and the current image because the sentence transformer typically does not have sufficient representational capacity for detailed reasoning. 
In turn, we feed the $m$ texts, as well as the user prompt, as a single query to GPT-4 \cite{openai2023gpt4}, and ask GPT-4 to find the $m^*$ most relevant captions, where $m^* << m$. 
The initial coarse matching based on embedding vectors is necessary because GPT-4 has a prompt character limit and a text description of every gallery image does not fit within this limit.
The reference images associated with the $m^*$ captions are then provided to the user in Step 4. 
Examples of user queries, objects detected in the scene, and the resulting LLM reference image suggestions (with an explanation from the LLM) are shown in \Cref{fig:llm_query}.
We used $m^*=$ 3 and $m=$ 16 in this work.
We set $m^*=$ 1 to have a single suggestion visualized in \Cref{fig:llm_query} only. 

\subsection{Semantic Keypoint Correspondence} 
\label{subsec:semkpt}

Since the reference image is captured from a different scene and contains object instances with highly different appearances, traditional local, appearance-based features (e.g., SIFT \cite{lowe2004distinctive}) are inadequate for establishing correspondences. 
To address this challenge, we exploit recent advances in self-supervised vision transformers to extract high-level semantic correspondences between the current and reference views. 
We follow the approach of Amir et al.\ \cite{amir2021deep} to establish semantic correspondence between a reference image and the current view captured by our RGB-D camera.
The reference image is taken from a different scene, but it is assumed that the current scene and the reference image contain semantically-similar elements. 
To extract features, we feed the image into a pre-trained DINO-ViT transformer \cite{caron2021emerging} and use the keys from intermediate transformer layers as dense image descriptors. 
Each key can be interpreted as a descriptor for the image patch associated with the corresponding token. 
The intermediate layers have been shown to offer a good trade-off between semantic and position information, both of which are important for semantic keypoint matching \cite{amir2021deep}.
Additional context is added to each descriptor by aggregating descriptors from adjacent patches via logarithmic binning. 
Having extracted descriptors from both the reference image and the RGB channels of the current view, we identify Best-Buddies Pairs (BBPs) \cite{oron2017best} between the descriptors from the two views to find correspondences.

Next, we select $k$ salient and well-distributed correspondences for use when solving the PnP problem. 
To achieve this, we concatenate the descriptors of each correspondence pair and run K-means on the concatenated descriptors with $k$ as the number of clusters. 
Finally, we select the single most salient correspondence from each cluster to form the final set of $k$ correspondences. 
Saliency is computed by averaging CLS attention heads in the last layer of the transformer for the image patches associated with the keypoint pair. 
Since depth is available, we back-project the 2D keypoints to obtain 3D keypoints for the current view.

\subsection{Camera View Adjustment} 
\label{subsec:vpalign}
We formulate the view adjustment problem as a PnP problem. 
Given 3D keypoints from the current view and the corresponding 2D keypoints from the reference image, we can solve for the camera transformation that would adjust the current viewpoint to align with that of the reference image. 
Let $\mathbf{x}_i=[x_i, y_i, 1]$ be the 2D homogeneous coordinates of the \textit{i}-th keypoint in the reference image, and $\mathbf{X_i}=[X_i, Y_i, Z_i, 1]$ be the 3D homogeneous coordinates of the corresponding keypoint in the current view. 
The PnP problem is to find a 3D camera transformation, consisting of a rotation $\mathbf{R} \in \mathrm{SO}(3)$ and translation $\mathbf{t} \in \mathbb{R}^{3}$, such that the sum of squared reprojection errors $\varepsilon_i$, $i = 1, \dots, n$,
\begin{equation}\label{eq:pnp}
    \varepsilon_i = ||\mathbf{x_i} - \mathbf{K} [\mathbf{R} | \mathbf{t}]\mathbf{X_i}||_2,
\end{equation}
is minimized.
Here, $\mathbf{K}$ is the $3 \times 4$ intrinsic matrix of the camera that we control. 
The intrinsic parameters of the camera used to take the reference image are not required, and we assume that they are unknown. 
Our implementation uses the popular EPnP \cite{lepetit2009ep} method, which provides a closed-form solution given at least four correspondences.

The reference image typically has a different resolution and aspect ratio compared to those captured by our camera. 
Our proposed approach is to output a photo that has the same aspect ratio as the reference image and at the highest resolution possible. 
To achieve this, we scale the reference image to be as large as possible without changing its aspect ratio, subject to the constraint that its width and height do not exceed that of our captured image. 
Then, we pad the reference image as needed such that it has the same dimensions as the captured image.
After a photo is taken, we can simply crop out the padded region to obtain a final image with the same aspect ratio as the reference. 
\Cref{fig:sample_gallery} illustrates the padding added to the reference images and the final, cropped photos.
Intuitively, scaling the reference image in this way ensures that the PnP solver outputs a camera pose where the reference visual elements are centered and fill the field of view to the extent possible. 

\subsection{Outlier Removal} 
\label{subsec:outliers}

Typically, a subset of the feature correspondences are erroneous. 
Often, this is due to different parts of the scene having similar appearances. 
For our problem, spurious correspondences may also arise when different parts of the scene having similar semantic interpretations. 
In turn, PnP solvers are usually used in conjunction with a robust estimator such as RANSAC \cite{fischler1981random} to remove spurious matches (i.e., outliers) and produce a solution based on a subset of mutually-coherent matches (i.e., inliers) only. 
That is, \Cref{eq:pnp} should be solved using the inliers only.

In a typical RANSAC framework for PnP, minimal samples of three to four correspondences are drawn and used to estimate a candidate PnP solution. 
Then, all correspondences are considered based on the candidate solution: those that are consistent with the candidate solution are considered as inliers and the rest are considered as outliers. 
Finally, the PnP solution with the largest number of inliers is selected and further refined using all of the inliers.
Determining the consistency between correspondences and a candidate solution is a key aspect of RANSAC. 
Usually, a correspondence $j$ is considered to be an inlier if the reprojection error $\varepsilon_j$ falls below some threshold $\tau$, which must be carefully tuned. 
Most applications of PnP assume that all corresponding points are captured from the same scene, and so it is often sufficient to fix $\tau$ to be a small value (e.g., 10 pixels).
In our problem setting, however, the 2D points from the reference image are not captured from the same scene as the 3D points (which come from from the camera's current view).
Additionally, a person may pose more similarly to one reference image than another.
Since the degree of discrepancy between the current scene and the reference image varies between problem instances, standard methods for filtering out spurious correspondences are difficult to apply. 
In \Cref{sec:exp}, we show that the optimal RANSAC threshold changes from one problem instance to another.

To address the challenges associated with threshold selection, numerous robust estimators have been developed that reduce the need for threshold tuning \cite{stewart1995minpran,torr2000mlesac,moisan2004probabilistic, barath2020magsac++}.
We use MAGSAC++ \cite{barath2020magsac++}, a state-of-the-art robust estimator that does away with the hard inlier threshold. 
For our PnP problem, MAGSAC++ solves a weighted least squares problem using all correspondences to minimize \Cref{eq:pnp}.
The weight of each correspondence is based on the expected likelihood of it being an inlier when marginalized over all threshold levels up to a maximum threshold. 
In this way, MAGSAC++ only requires specifying a maximum threshold, which can be loosely set to a large value.

In practice, we find it useful to have the system make multiple successive camera adjustments before taking the final photo. 
As the camera approaches the correct viewpoint, the quality of keypoint matches tends to improve, and the number of inliers tends to increase. 
Thus, the PnP solution will typically improve beyond the initial camera adjustment. 
In our implementation, we terminate the process when the mean distance in pixels between the $k$ keypoint correspondences does not improve for two iterations.

\section{Experiments} \label{sec:exp}
\begin{figure}
    \includegraphics[width=\columnwidth]{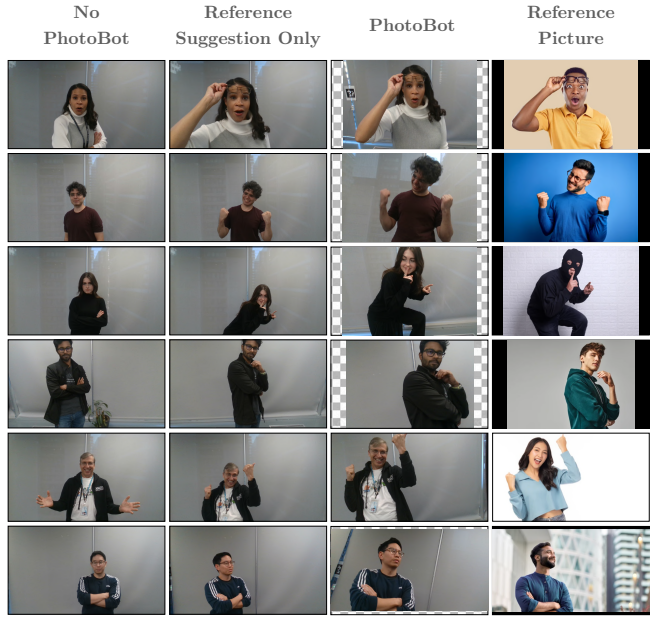}
    \centering
    \caption{
    Sample photos of users evoking various emotions. The user prompts, from top to bottom, are \emph{surprised}, \emph{confident}, \emph{guilty}, \emph{confident}, \emph{happy}, and \emph{confident}. Columns, from left to right, are: user's own creative posing; user mimicking the suggested reference using a static camera; photo taken by our PhotoBot system; and reference image suggested by PhotoBot. The checkered background indicates cropping. The black background indicates padding of the reference image to facilitate the PnP solution. PhotoBot automatically crops the photos it takes to match the image template.
    } 
	\label{fig:sample_gallery}
\end{figure}

We evaluated PhotoBot using a real Franka Emika robot manipulator equipped with a RealSense D435 RGB-D camera. 
We deployed PhotoBot and took photos of various scenes involving both humans and objects. 
First, we conducted a user study to evaluate the effectiveness of our reference suggestions and the view adjustment procedure. 
Second, we studied the effects of the crucial RANSAC inlier reprojection error threshold $\tau$ on the quality of the PnP solution when using DINO-ViT features.   
Third, we investigated the quality of the solutions as a function of the number of keypoints $k$.
Finally, we qualitatively evaluated whether PhotoBot is able to generalize to reference images with significantly larger distribution shifts (e.g., paintings).

\subsection{Human Preference Evaluation}
\begin{figure}
  \centering
  \begin{subfigure}[b]{.49\linewidth}
    \centering
    \includegraphics[width=\linewidth]{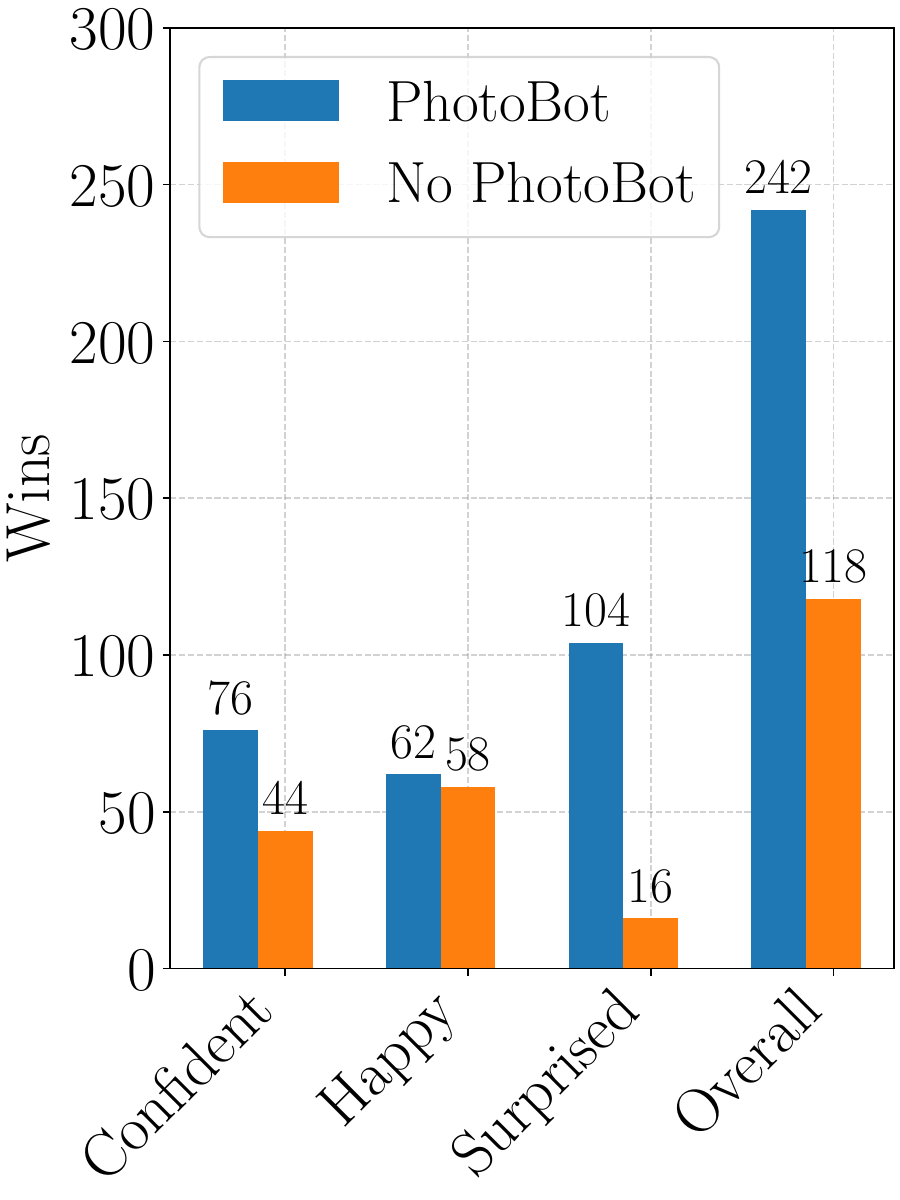}
    \subcaption{User preference study.}
    \label{subfig:template_suggestion}
  \end{subfigure}
  \hfill
  \begin{subfigure}[b]{.49\linewidth}
    \centering
    \includegraphics[width=\linewidth]{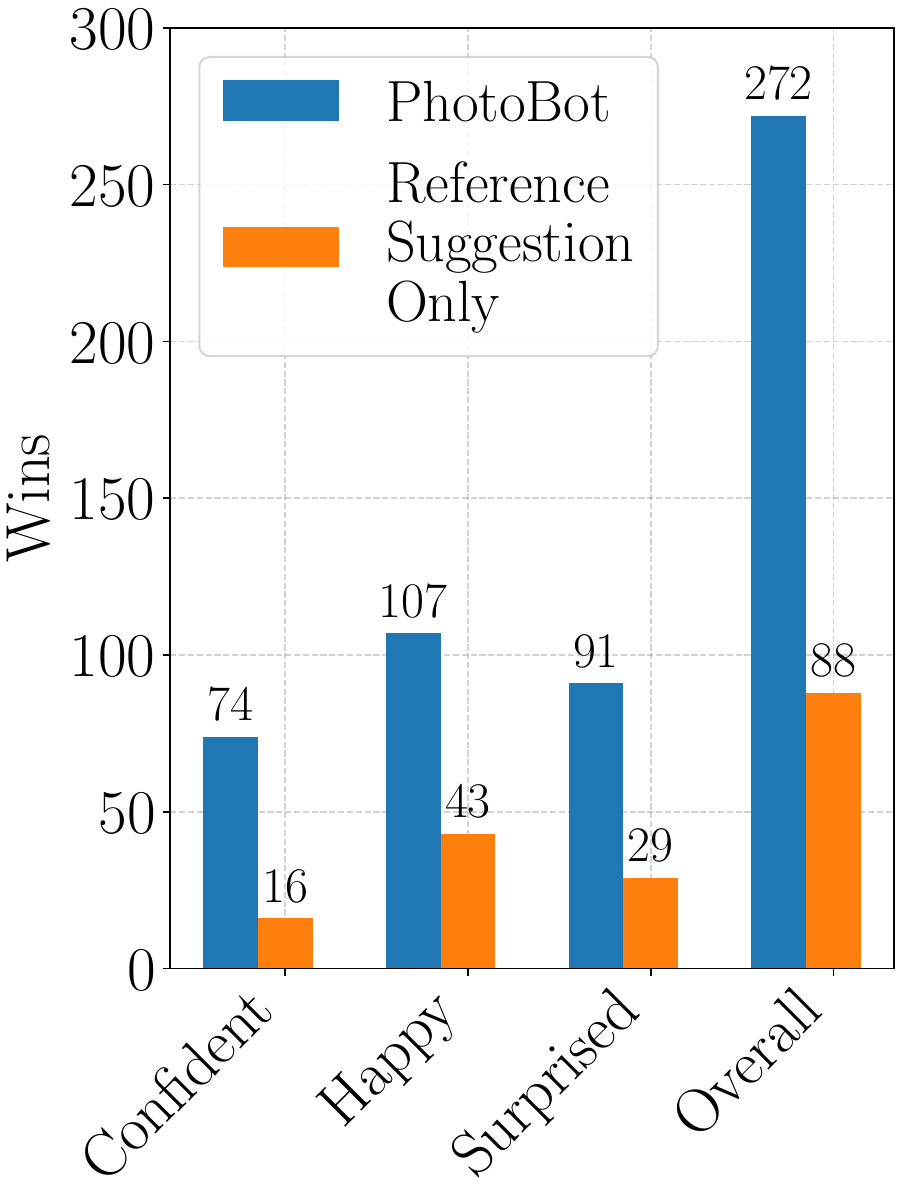}
    \subcaption{View adjustment evaluation.}
    \label{subfig:view_alignment}
  \end{subfigure}
  \caption{Results from our user studies. (a) User preference study based on the overall aesthetic quality and how well the user prompt is addressed by photos taken by PhotoBot, compared to the No PhotoBot baseline. (b) View adjustment evaluation based on how close the photos taken by PhotoBot and the Reference Suggestion Only baseline are to the actual reference image, in terms of the camera viewing angle and the photo layout. We present results for each category of emotion, and the aggregated value combining all emotions.}
  \label{fig:methodoverview}
\end{figure}
\begin{figure*}[]
  \centering
  \begin{subfigure}[t]{0.325\textwidth}
    \centering
    \includegraphics[height=0.75in]{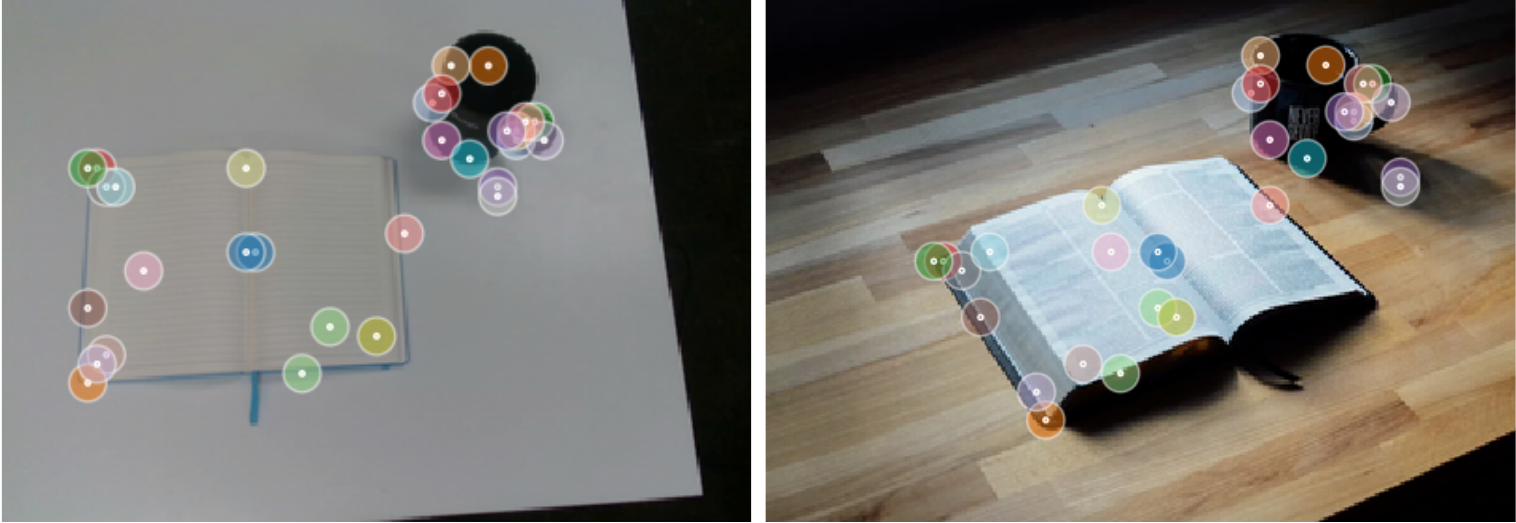}
    \subcaption{Mug and Book.}
    \label{subfig:mug_and_book_scene}
  \end{subfigure} \hfil
  \begin{subfigure}[t]{0.325\textwidth}
    \centering
    \includegraphics[height=0.75in]{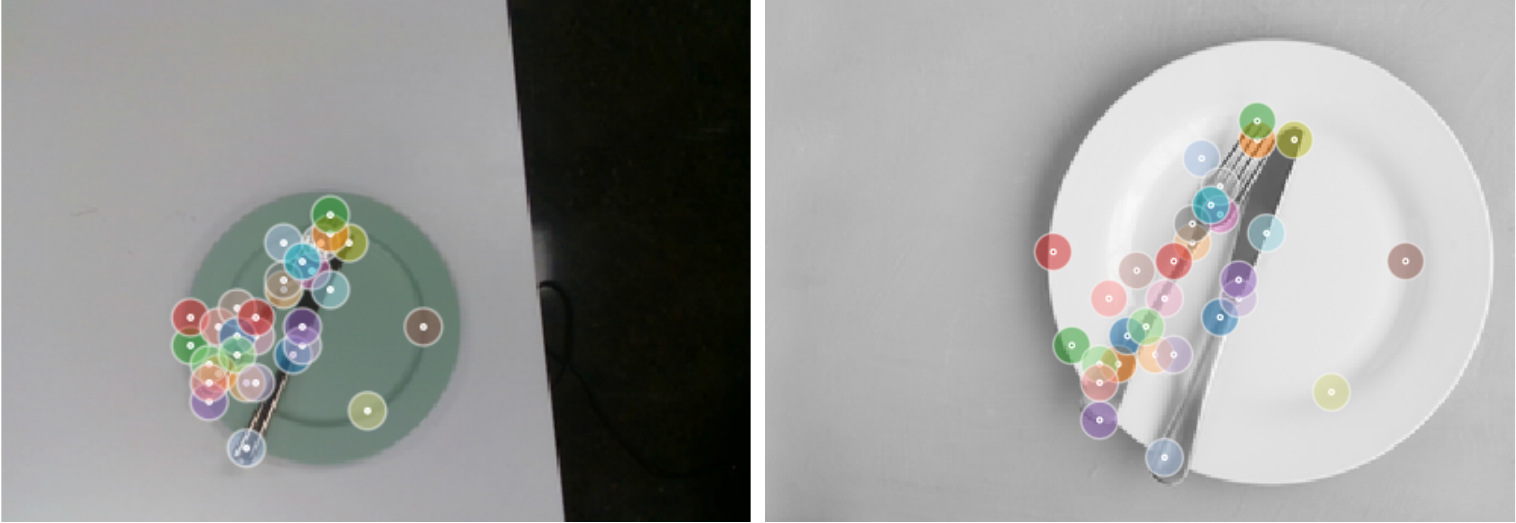}
    \subcaption{Plate and Utensils.}
    \label{subfig:plate_and_utensils_scene}
  \end{subfigure} \hfil
  \begin{subfigure}[t]{0.325\textwidth}
    \centering
    \includegraphics[height=0.75in]{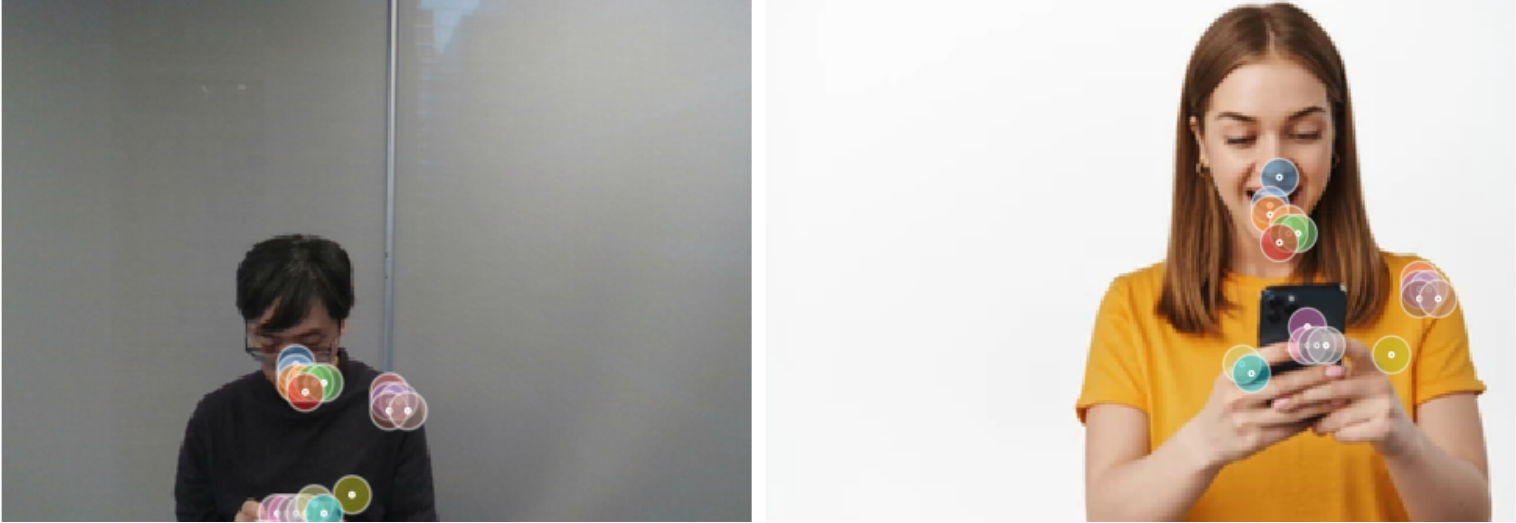}
    \subcaption{Human and Cellphone.}
    \label{subfig:cellphone_scene}
  \end{subfigure}\\
  \vspace{1.8mm}
  \begin{subfigure}[t]{0.325\textwidth}
    \centering
    \includegraphics[height=0.75in]{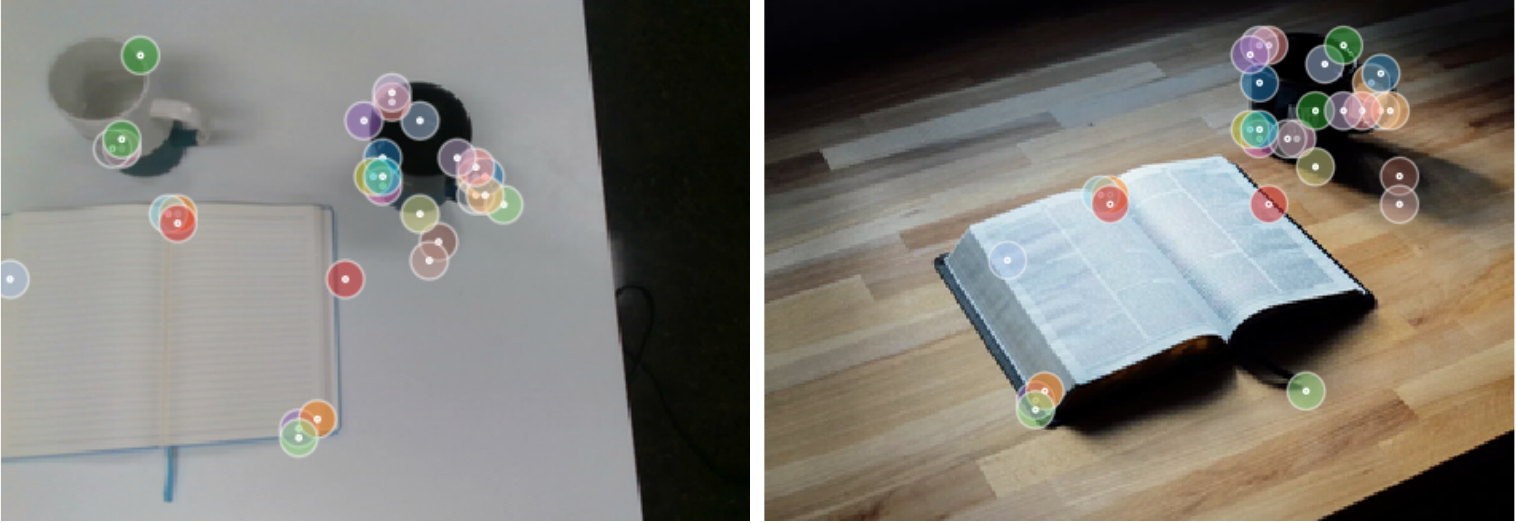}
    \subcaption{Mug and Book with Distractions.}
    \label{subfig:mug_and_book_outlier_scene}
  \end{subfigure} \hfil
  \begin{subfigure}[t]{0.325\textwidth}
    \centering
    \includegraphics[height=0.75in]{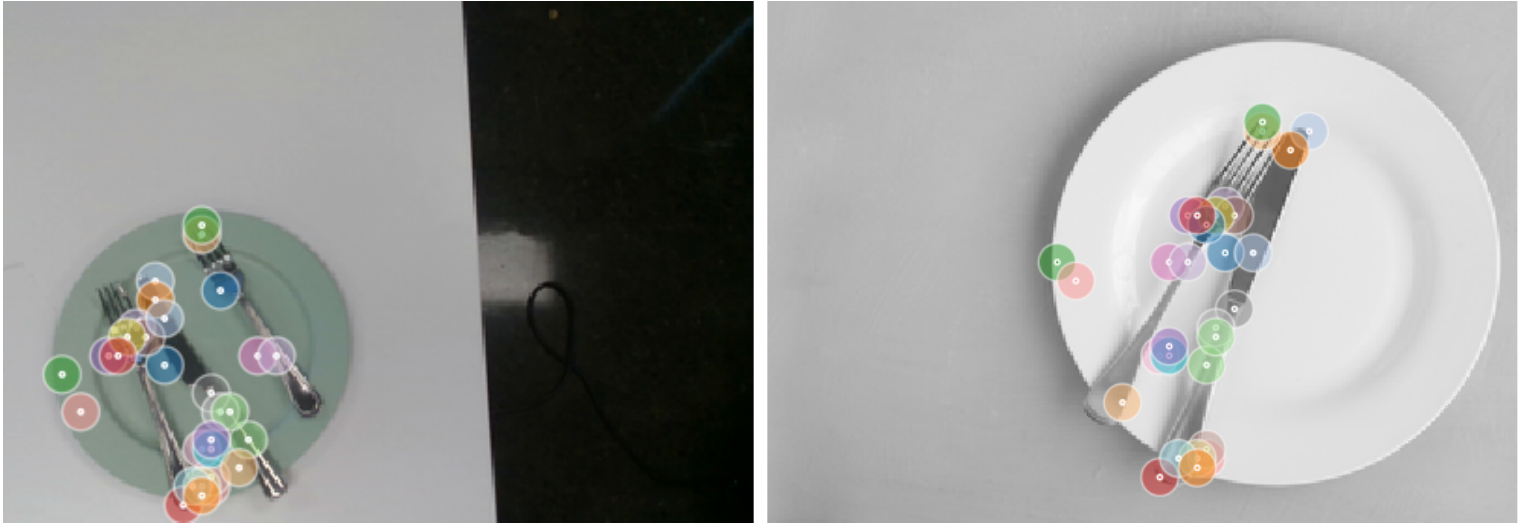}
    \subcaption{Plate and Utensils with Distractions.}
    \label{subfig:plate_and_utensils_outlier_scene}
  \end{subfigure} \hfil
   \begin{subfigure}[t]{0.325\textwidth}
    \centering
    \includegraphics[height=0.75in]{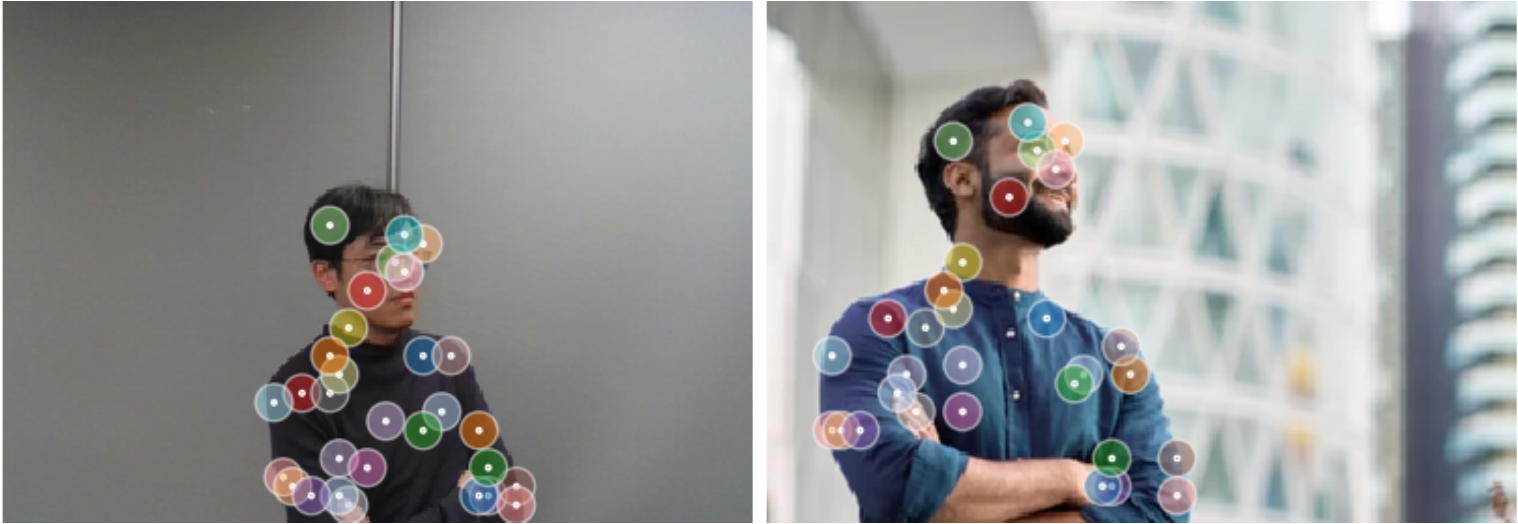}
    \subcaption{Confident Human.}
    \label{subfig:confident_scene}
  \end{subfigure}
  \caption{Scenes used to analyze and evaluate RANSAC inlier threshold with the learned semantic-level keypoints (i.e., DINO-ViT) used in the PhotoBot view adjustment procedure. For each pair, an initial image captured by PhotoBot is shown on the left and the reference image is shown on the right. We identify the point correspondences between image pairs with matching colours. }
  \label{fig:ransac_experiment_scenes}
  \vspace{-3mm}
\end{figure*}

To critically assess our reference-based photography approach, we had a group of users ($N =$ 8) interact with and have their photos taken by PhotoBot. 
We prompted users to query PhotoBot for image suggestions from three general categories of emotions: \textit{confident}, \textit{happy}, and \textit{surprised}.   
Based on each query, three reference images suggestions were provided, using the procedure detailed in \Cref{subsec:ts}, from which the user chose one.
For this experiment, we used a gallery of 75 images in total with images ranging from a variety of emotions.
Users then posed in a manner similar to the reference image (to the best of their ability) and PhotoBot took their photos while executing the camera view adjustment procedure outlined in \Cref{subsec:vpalign}.

We first investigated whether PhotoBot takes aesthetically-pleasing photos that satisfy (or match) the query from the user.
As a baseline, which we name ``No PhotoBot," we asked the same set of users to come up with a gesture and expression on their own that matched the category of emotion and to pose for a photo in front of a static camera.
Users took the ``No PhotoBot" photos before interacting with PhotoBot so that they were not influenced by any reference images.
Similar to taking a `selfie,' we allowed users to view the image from the fixed camera as they posed.
Examples of ``No PhotoBot" are shown in the first column of \Cref{fig:sample_gallery}.
We then surveyed a separate group of individuals ($M =$ 20), unrelated to this work, to evaluate the aesthetic quality of the photos from PhotoBot and ``No PhotoBot.''
We simultaneously presented the surveyed users with two photos, one from PhotoBot and one from ``No PhotoBot,'' and asked them to pick the photo that was both (1) more aesthetically pleasing and (2) better addressed the query from the user. 
We visualize the results from the survey in \Cref{subfig:template_suggestion}, separated by the categories of emotions, as well as the aggregated result over all three categories.
A vote for a photo consisted of a ``win'' for that particular method.
PhotoBot significantly outperforms the ``No PhotoBot" baseline in two out of three categories of emotions, \textit{confident} and \textit{surprised} and performs nearly on par for the \textit{happy} category.
Although aesthetics is subjective, our results show that the combination of reference suggestion and view adjustment from PhotoBot generally leads to more aesthetically-pleasing photos that better address a user query.

As a second experiment, we explored whether the PhotoBot view adjustment procedure leads to photos that better match the reference images. 
We use a baseline called ``Reference Suggestion Only,'' where we asked users to try their best to position themselves and pose in front of the static camera in a manner that re-created the reference image as closely as possible.
Examples of ``Reference Suggestion Only'' are shown in the second column of \Cref{fig:sample_gallery}.
For the PhotoBot picture, we used the ``Reference Suggestion Only'' pose from the user as the initial image for view adjustment.
We surveyed the same 20 individuals as before by simultaneously presenting a photo from PhotoBot, a photo from the ``Reference Suggestion Only'' baseline, and the reference image itself.
We then asked each individual to choose the photo that best matched the reference image in terms of the viewing angle and the layout. 
We visualize the results from the survey in \Cref{subfig:view_alignment}, separated by the specific categories of emotions, as well as the aggregated result over all three categories.
PhotoBot outperforms the ``Reference Suggestion Only" baseline in all three categories.

\subsection{RANSAC Inlier Threshold Evaluation} \label{sec:ransac_eval}
\begin{figure*}[]
  \centering
  \hspace*{1em}\begin{subfigure}[t]{0.30\textwidth}
    \centering
    \includegraphics[height=1in]{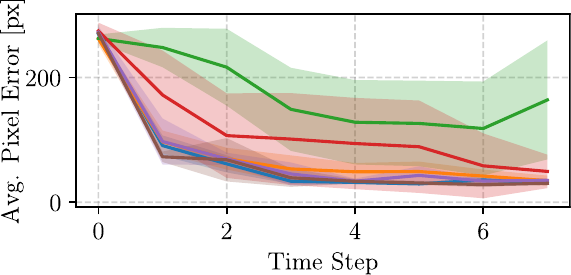}
    \vspace{-5mm}
    \subcaption{Mug and Book.}
    \label{subfig:mug_and_book}
  \end{subfigure} \hfill
   \begin{subfigure}[t]{0.30\textwidth}
    \centering
    \includegraphics[height=1in]{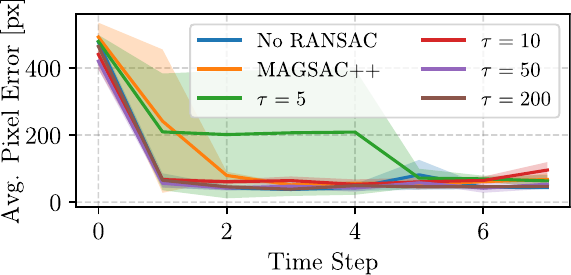}
    \vspace{-5mm}
    \subcaption{Plate and Utensils.}
    \label{subfig:plate_and_utensils}
  \end{subfigure} \hfill
   \begin{subfigure}[t]{0.30\textwidth}
    \centering
    \includegraphics[height=1in]{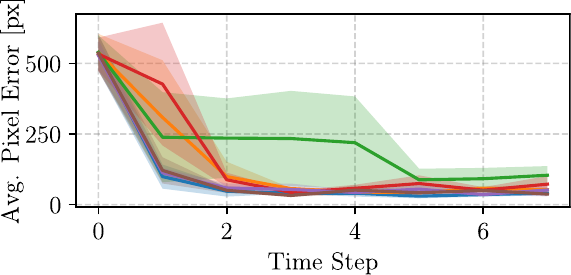}
    \vspace{-5mm}
    \subcaption{Human and Cellphone}
    \label{subfig:cellphone}
  \end{subfigure}%
  \hspace{3mm}
  \smallskip
  \vspace{1mm}
  \hspace*{1em}\begin{subfigure}[t]{0.30\textwidth}
    \centering
    \includegraphics[height=1in]{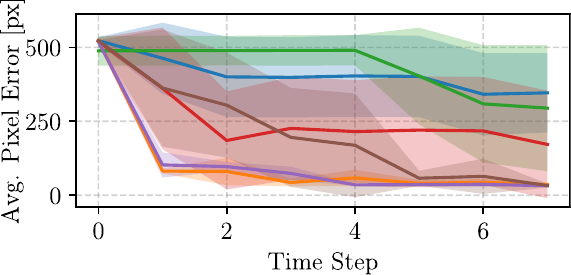}
    \vspace{-5mm}
    \subcaption{Mug and Book with Distractions.}
    \label{subfig:mug_and_book_outlier}
  \end{subfigure} \hfill
  \begin{subfigure}[t]{0.30\textwidth}
    \centering
    \includegraphics[height=1in]{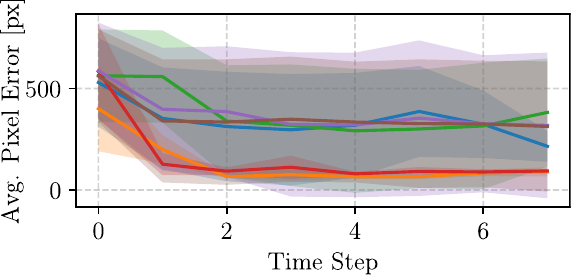}
    \vspace{-5mm}
    \subcaption{Plate and Utensils with Distractions.}
    \label{subfig:plate_and_utensils_outlier}
  \end{subfigure} \hfill
   \begin{subfigure}[t]{0.30\textwidth}
    \centering
    \includegraphics[height=1in]{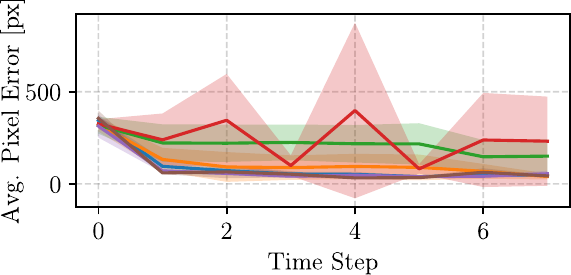}
    \vspace{-5mm}
    \subcaption{Confident Human}
    \label{subfig:confident_human}
  \end{subfigure}%
  \hspace{3mm}
  \vspace{1mm}
  \caption{Average absolute pixel errors with various RANSAC methods for four different template images and scenes. We measure average pixel error over all keypoints for eight time steps of view adjustment. The shaded region consists of one standard deviation measured over 3 repeated runs. The best-performing inlier reprojection error threshold $\tau$ varies between scenes. The adaptive MAGSAC++ algorithm performs close to the best fixed-threshold parameter with RANSAC.}
  \label{fig:ransac_error_curves}
  \vspace{-3mm}
\end{figure*}

\begin{figure}[b!]
	\vspace{-3.5mm}
    \includegraphics[height=3.75cm]{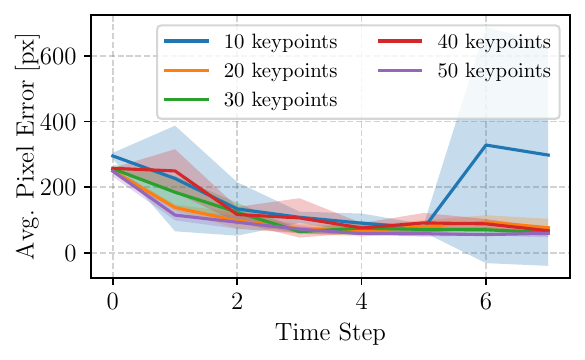}
    \centering
    \caption{Pixel errors at each time step for various numbers of DINO-ViT keypoints used. The shaded region consists of one standard deviation measured over three repeated runs.} 
	\label{fig:num_pairs}
\end{figure}

We experimentally evaluated and analyzed the effect of the RANSAC inlier reprojection error threshold $\tau$ on the quality of PnP solutions when using DINO-ViT keypoints.
As previously highlighted in \Cref{sec:me}, finding semantic keypoint correspondences between different scenes poses an interesting challenge when compared to finding correspondences between local appearance-based features of the same scene. 
In particular, the magnitude of reprojection errors and the number of outliers can vary significantly across different scenes and users may pose at varying levels of similarity to the reference image.

We considered the scenes shown in \Cref{fig:ransac_experiment_scenes} in our experiments. 
We chose four settings consisting of static objects (``Mug and Book,'' ``Mug and Book with Distractions,'' ``Plate and Utensils,'' and ``Plate and Utensils with Distractions'') to be able to test in a consistent and reproducible manner. 
For each set of objects, we included a version with and without distractor objects that introduced additional outliers.
Distractor objects are repeated objects in the scene, not present in the template image (e.g., an extra mug or fork), that induce false correspondences. 
Additionally, we considered two scenes involving humans (``Human and Cellphone'' and ``Confident Human'').
For each reference image, we re-created the scene with similar objects or with a human model and then ran our view adjustment procedure.
We tested a total of six methods: RANSAC with four different fixed threshold pixel values of $\tau = \{$5, 10, 50, 200$\}$, No RANSAC, and MAGSAC++ with a maximum error threshold of 50 pixels.
For each method, we ran view adjustment for eight successive steps, repeating the experiment three times for each scene. 
At each step, if the PnP solver did not find a solution or the suggested pose was outside the workspace of the robot, then the robot did not move for that step.

\Cref{fig:ransac_error_curves} displays the results for each respective scene and reference image. 
We measured the error between the reference image and the current image by manually annotating the ground truth correspondences at each step. 
During annotation, it was up to the human annotator to select keypoints that were salient and meaningful (e.g., the tip of the fork or the handle of the mug).
We observed that the best-performing fixed threshold differs for each scene.
Furthermore, for DINO-ViT features, larger error thresholds are required for PnP to produce solutions that converge.
In all scenes, setting a threshold value of $\tau=$ 5 caused the PnP solution to fail to converge frequently, which increased the overall error for the trial.
For all of the tested scenes, the adaptive MAGSAC++ algorithm performs similarly to the best variant of RANSAC with a fixed threshold. 

\begin{figure*}
  \begin{subfigure}[t]{0.35\textwidth}
    \centering
    \includegraphics[height=0.75in]{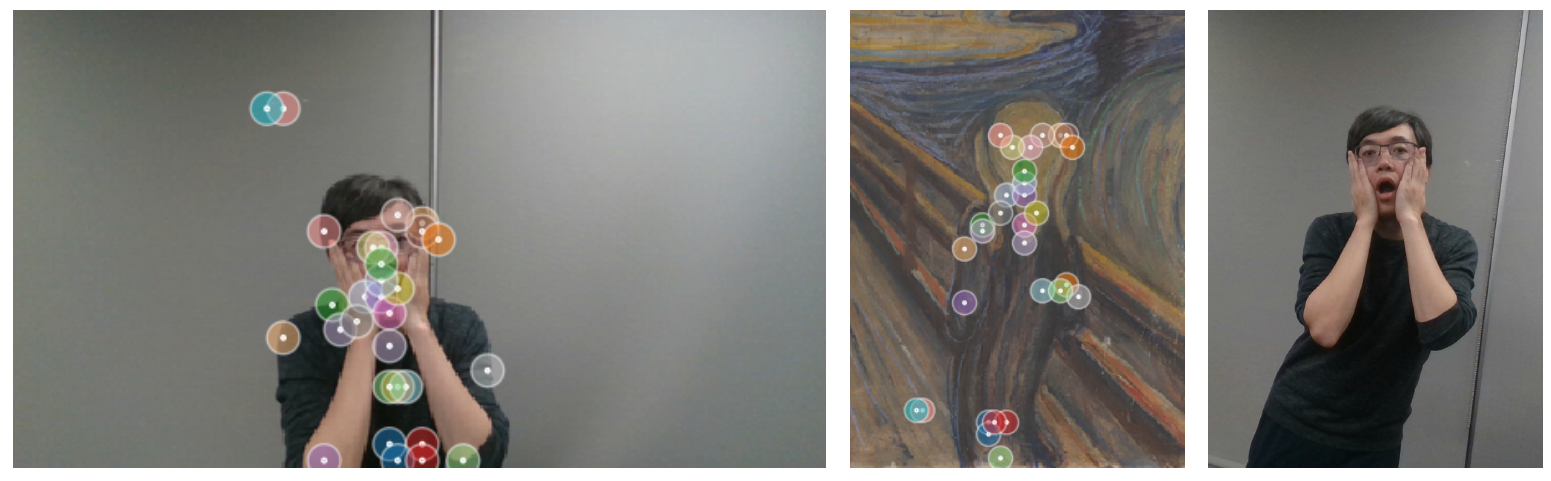}
    \subcaption{Using ``The Scream" as an image template.}
    \label{subfig:scream}
  \end{subfigure}
  \hfill
  \begin{subfigure}[t]{0.635\textwidth}
    \centering
    \includegraphics[height=0.7625in]{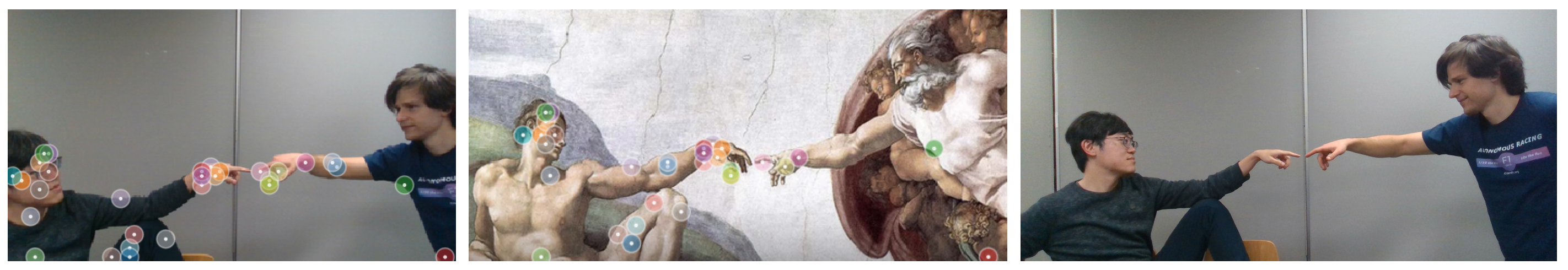}
    \subcaption{Using ``The Creation of Adam" as an image template.}
    \label{subfig:creation_of_adam}
  \end{subfigure} 
  \caption{PhotoBot can generalize to paintings used as reference images. For each image set above, the first and second images (left to right) show the initial set of correspondences found, while the third image is the photo with the highest-scoring alignment found by PhotoBot. The large distribution shift between the images introduces additional outliers.}
  \label{fig:ood_picture_template}
  \vspace{-5mm}
\end{figure*}

\subsection{Required Number of Keypoints Evaluation}

We evaluated the quality (i.e., average pixel error as measured in \Cref{sec:ransac_eval}) of the camera view adjustment procedure of PhotoBot as a function of the number of DINO-ViT keypoints. 
As previously discussed in \Cref{subsec:semkpt}, the number of keypoints $k$ is a hyperparameter that is set based on the number of chosen K-means clusters.
We ran PhotoBot on the Mug and Book scene shown in \Cref{subfig:mug_and_book} while varying the number of keypoints used, following the same experimental procedure in \Cref{sec:ransac_eval}. 
The results are visualized in \Cref{fig:num_pairs}.
Interestingly, we observed that PhotoBot performs reasonably well when at least 20 keypoints are used. 
The variant with 10 keypoints often diverged due to insufficient keypoint coverage across the entire scene.

\subsection{Generalizing to Other Reference Images}

We also carried out an evaluation with paintings, rather than photographs, to determine if PhotoBot is capable of generalizing to references from different mediums.
Notably, DINO-ViT features have been shown to generalize across large distribution shifts \cite{caron2021emerging, amir2021deep}. 
In \Cref{fig:ood_picture_template}, we show results using two famous paintings as references. 
Qualitatively, we observe that, despite the changes in medium and format, PhotoBot is still capable of finding snapshots that resemble the paintings. 
The large distribution shift introduces additional outliers, which MAGSAC++ is able to filter out.

\section{Conclusion}
\label{sec:con}

In this paper, we presented PhotoBot, a novel interactive photography assistant. 
PhotoBot is capable of suggesting reference images based on a natural language query from a user and a visual observation of the current scene.
In addition, PhotoBot can propose camera adjustments to match the layout and composition of a chosen reference image.
We conducted experiments demonstrating that the photos taken by PhotoBot are aesthetically pleasing, address the users' prompts, and follow the reference images' layouts and compositions.
We also studied various factors that affect the quality of the photos taken and showed that PhotoBot is able to generalize to other references sources, including paintings.
As future work, we aim to study alternate physical embodiments with a wider range of motion (e.g., a quadcopter or a mobile manipulator), and to develop methods to provide language-based corrective posing feedback to the user.

\section*{Acknowledgements}

All reference images are used under license from Shutterstock.com.
Individuals in the photographs gave consent for their images to appear in this paper.

\bibliographystyle{ieeetr}
\bibliography{refs}

\end{document}